\documentclass[final,3p,times]{elsarticle}

\usepackage{amssymb}
\usepackage{bm}
\usepackage{amsmath}
\usepackage[ruled]{algorithm2e}
\usepackage[usenames,dvipsnames]{color}
\usepackage{array}
\usepackage{subcaption}
\usepackage{graphicx}
\usepackage{tikz}
\usepackage{booktabs}
\usepackage{float}
\usepackage{caption}
\usepackage{placeins}
\usepackage{longtable}
\usepackage{tcolorbox}
\usepackage{tabularx}
\usepackage{enumitem}

\usetikzlibrary{shapes, arrows.meta, positioning, calc}
\graphicspath{{figs/}}
\newcommand{\doi}[1]{\href{https://doi.org/#1}{\nolinkurl{#1}}}

\usepackage{hyperref}
\hypersetup{ colorlinks, linkcolor={magenta}, citecolor={blue}, urlcolor={blue!80!black}, breaklinks=true, plainpages=true }

\newcommand{\cref}[3]{\hyperref[#2]{#1~\ref*{#2}{#3}}}
\newcommand{\crefs}[3]{\hyperref[#2]{#1~\ref*{#2}-\ref*{#3}}}
\newcommand{\colref}[2]{\hyperref[#2]{#1~\ref*{#2}}}

\newcommand{\figref}[1]{\colref{Figure}{#1}}

\newcommand{\figsref}[2]{\crefs{Figures}{#1}{#2}}
\newcommand{\secref}[1]{\colref{Section}{#1}}

\newcommand{\tabref}[1]{\colref{Table}{#1}}

\begin{document}
\journal{Computers and Electronics in Agriculture}

\begin{frontmatter}

\title{AI-enabled Low-Cost 3D Maize Ear Morphometry Platform at Breeding Scale}

\author[1]{Therin Young\texorpdfstring{\fnref{eq}}{}}

\author[2]{Elijah Rodriguez\texorpdfstring{\fnref{eq}}{}}

\author[3]{Lisa Coffey}

\author[2]{Talukder Zaki Jubery}

\author[1,2]{Adarsh Krishnamurthy}

\author[3]{Patrick Schnable\texorpdfstring{\corref{cor1}}{}}
\ead{schnable@iastate.edu}

\author[1,2]{Baskar Ganapathysubramanian\texorpdfstring{\corref{cor1}}{}}
\ead{baskar@iastate.edu}
\cortext[cor1]{Corresponding authors}

\fntext[eq]{These authors contributed equally to this work.}

\affiliation[1]{organization={Department of Mechanical Engineering},
addressline={Iowa State University}, city={Ames}, postcode={50011}, state={Iowa}, country={USA}}

\affiliation[2]{organization={Translational AI Research and Education Center},
addressline={Iowa State University}, city={Ames}, postcode={50011}, state={Iowa}, country={USA}}

\affiliation[3]{organization={Department of Agronomy},
addressline={Iowa State University}, city={Ames}, postcode={50011}, state={Iowa}, country={USA}}

\begin{abstract}
Maize ear geometry (length, width, curvature, and volume) is closely tied to yield and grain-filling outcomes, but existing high-throughput phenotyping pipelines remain constrained by the cost, labor, and specialized hardware they require. We developed and validated a low-cost pipeline that reconstructs a watertight 3-D mesh of a maize ear from a single 20-second video captured with a consumer-grade DSLR on a motorized turntable under uniform LED illumination. Camera poses from a multi-seed COLMAP procedure initialize a Neural Radiance Field (NeRF), and a cylindrical holder of known diameter, visible in every frame, provides automatic metric scaling with downstream geometric quality control. Applied to 300 ears spanning a diverse maize inbred panel, 250 (83.3\,\%) passed automated processing and quality control. Skeleton length agreed with manual caliper measurements across all 250 ears ($R^{2} = 0.964$, RMSE $= 4.68$\,mm), and convex-hull volume agreed with water-displacement volume on a 15-ear subset spanning the full size range ($R^{2} = 0.982$, RMSE $= 5.26$\,mL). Residual length error grew with ear curvature, whereas bounding-box height, which records the same straight-line chord as calipers, showed no such trend; the discrepancy therefore originates in the measurement definition, since calipers record the chord while skeleton length traces the geodesic arc. The capture hardware costs approximately 607 USD, and operator involvement fell from roughly five minutes to one minute per ear, with all downstream processing running unattended. The platform provides a foundation for breeding-scale 3-D ear phenotyping.
\end{abstract}

\begin{graphicalabstract}
\includegraphics[width=1\linewidth]{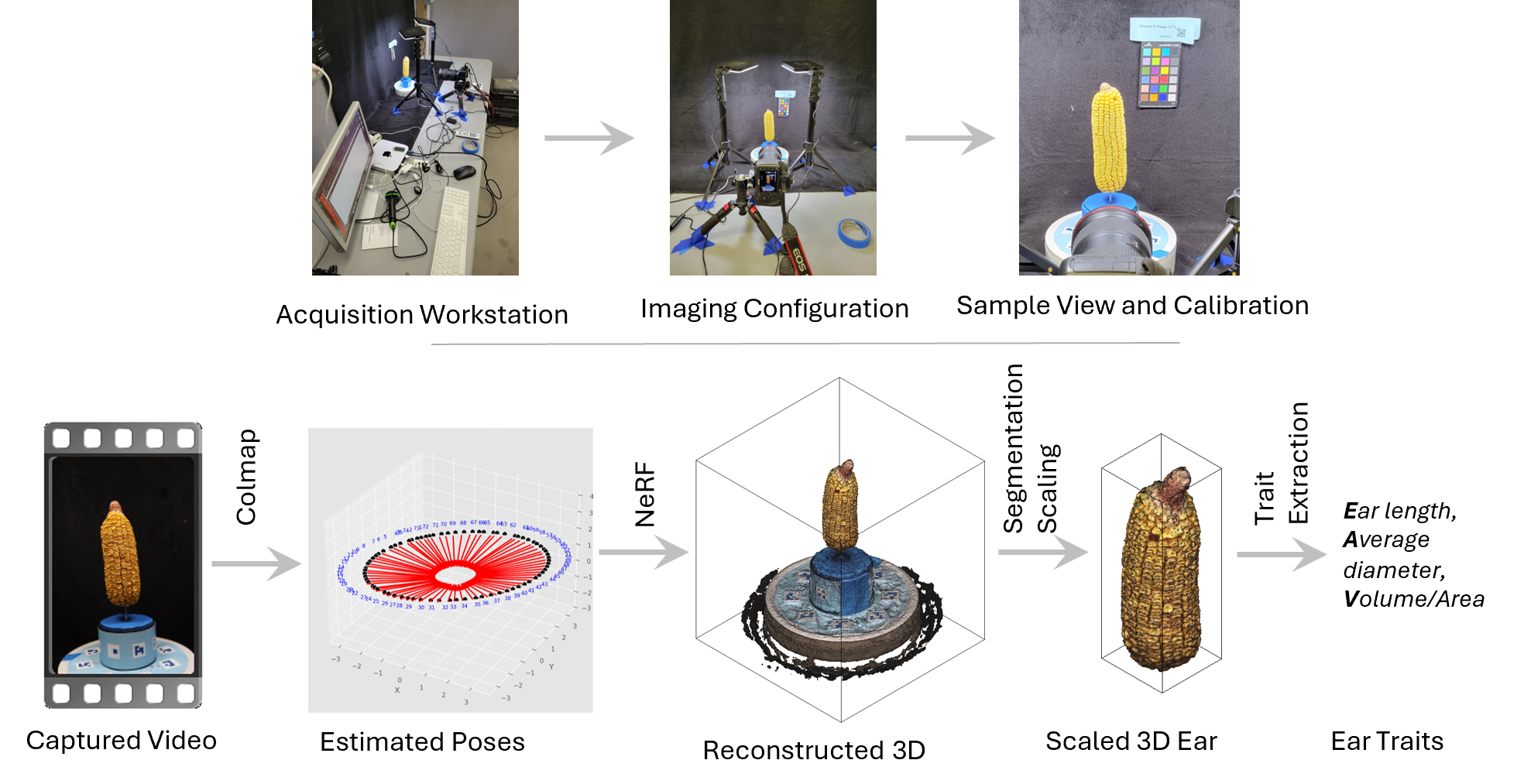}
\end{graphicalabstract}

\begin{highlights}
   \item Low-cost stationary-camera NeRF pipeline gives 3-D maize ear traits from video.
   \item Ten-seed COLMAP with intrinsic averaging enforces 100\% frame registration.
   \item A known-diameter holder scales the mesh; quality control flags failed fits.
   \item One command runs frames, COLMAP, NeRF, scaling and trait output to CSV.
   \item Length and volume match calipers and displacement across 250 maize ears.
\end{highlights}


\end{frontmatter}


\section{Introduction}

High-resolution characterization of maize ear traits such as total kernel number, row architecture, section-wise diameter profile, and overall shape is central to genetic analyses and yield prediction. However, the manual processes required to capture these traits are labor-intensive, subjective, and difficult to scale up for larger phenotyping studies that introduce additional constraints such as lighting, weather, and labor availability. Maize ear phenotyping is a technical challenge for high-throughput phenotyping research studies. Furthermore, the variation in maize ear geometries requires robust phenotyping pipelines capable of capturing these geometric and physical complexities.

Traditional two-dimensional (2-D) studies have produced low-cost, high-precision estimations of ear and cob length that are highly correlated with manual hand measurements \citep{Makanza2018-ho, Miller2017-xm}. Open-source projects like \textit{EarCV} leveraged fundamental computer vision principles to extract ear length and width, in addition to more complex traits like taper, curvature, tip fill, and color in diverse inbred datasets to assess each trait's relationship with yield \citep{Gonzalez2022-fd}. The open-source project \textit{EARBOX} adds to the literature by also accurately capturing ear diameter along the length of the ear to increase throughput and more accurately estimate spatial organization metrics from rotating ears \citep{Oury2021-zt}. However, pure 2-D methods cannot recover back-side kernels, true volumes, or complete 3-D spatial distributions and suffer from the usual limitations that are inherent of in-field studies, such as environmental occlusions, single-view imaging, and limited robustness to mixed textures and complex geometries. A more recent study by Fan et al. features a DIY maize ear-imaging platform with a deep learning-based end-to-end phenotypic data extraction pipeline \citep{Fan2026-lx}. Ear length, diameter, volume, and weight were calculated along with other grain-based traits which included kernel number (KN), kernel row number (KRN), kernel number per row (KPR), kernel thickness, kernel width, and thousand kernel weight. The imaging platform used in the study was very similar to that used in this study; a key distinction being that maize ear videos were not used for 3-D reconstructions. Rather, the videos were converted into projected 2-D images that represent a flattened or unrolled view of the maize ear which were then used as input features for training a CNN. Fan et al. demonstrate that video-based platforms can support both ear-level and kernel-level trait extraction, simultaneously. Kernel-level measurements are not yet directly extracted from the 3-D NeRF mesh in the current pipeline; the trait set reported here is therefore ear-level rather than kernel-level. A companion study \citep{Kumar2026-tppj} extends this same reconstruction platform to eleven ear- and kernel-level traits (including kernel count, kernel row number, kernels per row, per-kernel surface area, volume, packing geometry, hue, and aspect ratio) via cylindrical unwrapping of the point cloud and zero-shot kernel instance segmentation.
Three-dimensional (3-D) high-throughput phenotyping for maize ears can help address the limitations of 2-D studies by providing access to the full surface geometry of the ear, a prerequisite for volumetric traits and complete cross-sectional profiles that single-view imaging cannot capture. Volume, geodesic arc length, and cross-sectional eccentricity profiles have no counterpart in a single projection, so for these traits the relevant question is access rather than relative accuracy. The bottleneck lies in accurately recovering the ear surface and, for finer-grained analyses, segmenting structural components such as kernels, rows, and ear boundaries.

Classical multi-view stereo (MVS) approaches such as COLMAP \citep{Schoenberger2016_COLMAP,Schoenberger2016_MVS} and Metashape yield detailed surfaces but require well-controlled views, careful backgrounds, and expert tuning, and are slow; for commercial packages, licensing cost and operator effort limit routine use at breeding scale \citep{Wang2019_Maize3DCompare}. Turntable rigs built from consumer hardware have already brought metric 3-D reconstruction of individual fruits within reach of a modest budget \citep{Feldmann2022_FruitForm}. Active 3-D sensors (structured light, laser) provide precise geometry, but are costly, calibration-heavy, and sensitive to surface reflectance and lighting \citep{Qiu2018_PhenoSensors,Wang2019_Maize3DCompare}. Ear-focused 3-D pipelines underline the phenotypic value of true 3-D while still relying on resource-intensive stages that hinder scaling \citep{Yang2024_EarRot3D}.

Neural Radiance Fields (NeRF) reconstruct continuous volumetric scenes from images \citep{Mildenhall2020_NeRF}. Modern implementations such as Instant-NGP and Nerfstudio's \textit{nerfacto} accelerate training using multi-resolution hash encodings and practical training recipes \citep{Mueller2022_InstantNGP,Tancik2023_Nerfstudio}. A stationary-camera variant (rotating object, fixed camera) simplifies capture and supports more delicate or bulky optics. Recent agricultural work reports high-fidelity plant geometry from short videos with COLMAP-based poses, both in the field \citep{Arshad2024_PlantNeRF} and against commercial MVS baselines \citep{Hu2024_PlantNeRF}. Learning-based radiance/geometry fields such as \textit{3D Gaussian Splatting} provide an additional path for fast, photorealistic scene models in phenotyping contexts \citep{Kerbl2023_3DGS}.

As remote sensing technologies such as LIDAR (active sensing) and photogrammetry (passive sensing) have become more available and efficient over the past decade, several high-throughput 3-D phenotyping studies have emerged and advanced the field. Many of these studies focus on micro-level traits of the maize ear, such as kernel row count and kernels per row \citep{SUN2026111235}. Zhu et al. used a Laplace-based skeleton extraction algorithm to generate a 3-D point cloud skeleton followed by several sub-skeleton refinements and localized constraints to accurately segment maize ear components and accurately measure ear height, length, diameter, and the ratio of plant height to ear height \citep{Zhu-Chao-2021-wc}. The $R^2$ values for automated vs manual hand measurements were 0.97, 0.78, 0.73, and 0.96, respectively. Sun et al. used structured light reconstructions for KPR and KRN detections achieved through the Density-Based Spatial Clustering of Applications with Noise (DBSCAN) algorithm with a directional erosion strategy to improve robustness for adherent kernels \citep{SUN2026111235}. Sun et al.'s study resembles our study in how the 3-D reconstructions were achieved passively. However, their use of structured light resulted in a higher-resolution point cloud, which made KPR and KRN detections accessible. The use of NeRF in our study coupled with COLMAP currently supports structurally accurate reconstructions at the macroscopic level; direct 3-D kernel-level measurements from the NeRF mesh have not yet been achieved at this resolution. Related work from our own group has separately shown that zero-shot instance segmentation with the Segment Anything Model \citep{Kirillov2023_SAM} can recover kernel counts directly from 2-D ear images without task-specific training \citep{Zaremehrjerdi2025-sam}. The ear-level trait set reported in the present study is a deliberately preliminary scope: a companion study \citep{Kumar2026-tppj} extends the same reconstruction platform to direct 3-D kernel-level extraction, cylindrically unwrapping the calibrated point cloud and applying zero-shot instance segmentation to jointly recover eleven ear- and kernel-level traits.

\begin{table}[t!]
\centering
\small
\setlength{\tabcolsep}{3pt}
\caption{Comparative analysis of automated maize ear phenotyping methods. Abbreviations marked $^{a}$ are outputs of the cited method only; all others are defined in \tabref{tab:trait_definitions}. Cost is a qualitative tier (\$~=~consumer-grade hardware; \$\$~=~specialist hardware; \$\$\$~=~specialist/industrial-grade or resource-intensive hardware) inferred from each study's reported equipment, not a verified price.}
\label{tab:method_comparison}
\renewcommand{\arraystretch}{0.95}
\begin{tabularx}{\textwidth}{@{} >{\raggedright\arraybackslash}p{3.2cm} >{\raggedright\arraybackslash}p{2.6cm} >{\raggedright\arraybackslash}p{1.6cm} >{\centering\arraybackslash}p{1.2cm} >{\raggedright\arraybackslash}X @{}}
\toprule
\textbf{Method} & \textbf{Modality} & \textbf{Hardware} & \textbf{Cost} & \textbf{Outputs} \\
\midrule
\citet{Makanza2018-ho,Miller2017-xm}
& 2-D image & Consumer & \$ & Ear length$^{a}$, cob length$^{a}$ \\[2pt]
\citet{Gonzalez2022-fd} (EarCV)
& 2-D image & Consumer & \$ & Ear length$^{a}$, ear width$^{a}$, Taper$^{a}$, Curvature$^{a}$, Tip Fill$^{a}$, Color$^{a}$ \\[2pt]
\citet{Oury2021-zt} (EARBOX)
& 2-D rotating image & Consumer & \$ & Ear length$^{a}$, diameter profile$^{a}$ \\[2pt]
\citet{Fan2026-lx}
& 360$^\circ$ video (2-D unrolled) & Consumer & \$ & Ear length$^{a}$, diameter$^{a}$, volume$^{a}$, weight$^{a}$, KN$^{a}$, KRN$^{a}$, KPR$^{a}$, kernel thickness$^{a}$, kernel width$^{a}$, TKW$^{a}$ \\[2pt]
\citet{Yang2024_EarRot3D}
& 3-D rotational reconstruction & Specialist, resource-intensive & \$\$\$ & Ear 3-D geometry \\[2pt]
\citet{Zhu-Chao-2021-wc}
& 3-D point cloud (Laplace skeleton) & Specialist & \$\$ & Ear height$^{a}$, ear length, ear diameter, plant-height ratio$^{a}$ \\[2pt]
\citet{SUN2026111235}
& Structured light & Specialist & \$\$\$ & KRN$^{a}$, KPR$^{a}$ \\
\midrule
\textbf{Ours} &
\textbf{Stationary-camera 360$^\circ$ video (NeRF)} &
\textbf{Consumer} &
\textbf{\$} &
\textbf{\boldmath Skeleton Length, Max Width, Chord/Arc Ratio, Taper-to-Tip, Curvature Score, Mean Eccentricity, Convex-Hull Volume, Surface Area$^{b}$} \\
\bottomrule
\multicolumn{5}{l}{\footnotesize $^{a}$Output of cited method only; see respective reference for definition.}\\
\multicolumn{5}{l}{\footnotesize $^{b}$Ear-level outputs of this platform paper; a companion study \citep{Kumar2026-tppj} extends this}\\
\multicolumn{5}{l}{\footnotesize \phantom{$^{b}$}point cloud to eleven ear- and kernel-level traits (kernel count, row number, kernels per row,}\\
\multicolumn{5}{l}{\footnotesize \phantom{$^{b}$}per-kernel surface area, volume, packing, hue, and aspect ratio).}\\
\end{tabularx}
\end{table}

\tabref{tab:method_comparison} summarizes existing automated maize ear phenotyping methods and situates the present work among them. Relative to 2-D pipelines, the present pipeline recovers true 3-D surface geometry rather than a single-view projection; relative to classical MVS and active 3-D sensors, it uses only consumer-grade hardware and requires no specialist calibration equipment; and relative to prior 3-D ear-phenotyping studies, it is the first to combine a stationary-camera NeRF reconstruction with fully automated metric scaling and quality control at breeding scale.

We propose a frugal, reproducible maize-ear phenotyping pipeline integrating established components (COLMAP pose estimation, Nerfstudio's \textit{nerfacto} model, and convex-hull surface reconstruction) with three design contributions:
\begin{enumerate}[leftmargin=*, itemsep=2pt, topsep=3pt]
  \item A \textbf{multi-seed COLMAP strategy} that runs ten independent reconstructions and retains only 100\,\%-coverage solutions, averaging intrinsics across the retained runs.
  \item A \textbf{known-diameter cylindrical fiducial} for automatic metric scaling without external calibration objects.
  \item A \textbf{single-command automation layer} that wraps all stages from video to a trait CSV file.
\end{enumerate}

The rest of the paper is structured as follows. \secref{sec:methods} details the imaging hardware, data acquisition protocol, multi-seed COLMAP pose estimation, NeRF reconstruction, metric scaling, geometry quality control, and trait-extraction procedures. \secref{sec:results} reports COLMAP registration outcomes, geometric validation against manual caliper and water-displacement references, and population-level trait distributions across the SAM diversity panel. \secref{sec:discussion} interprets these results in light of the pipeline's practical impact, design advantages, biological relevance, and current limitations, and outlines directions for future work.

\section{Materials and Methods}
\label{sec:methods}

\begin{figure*}[t]
    \centering
    \includegraphics[width=1.0\textwidth]{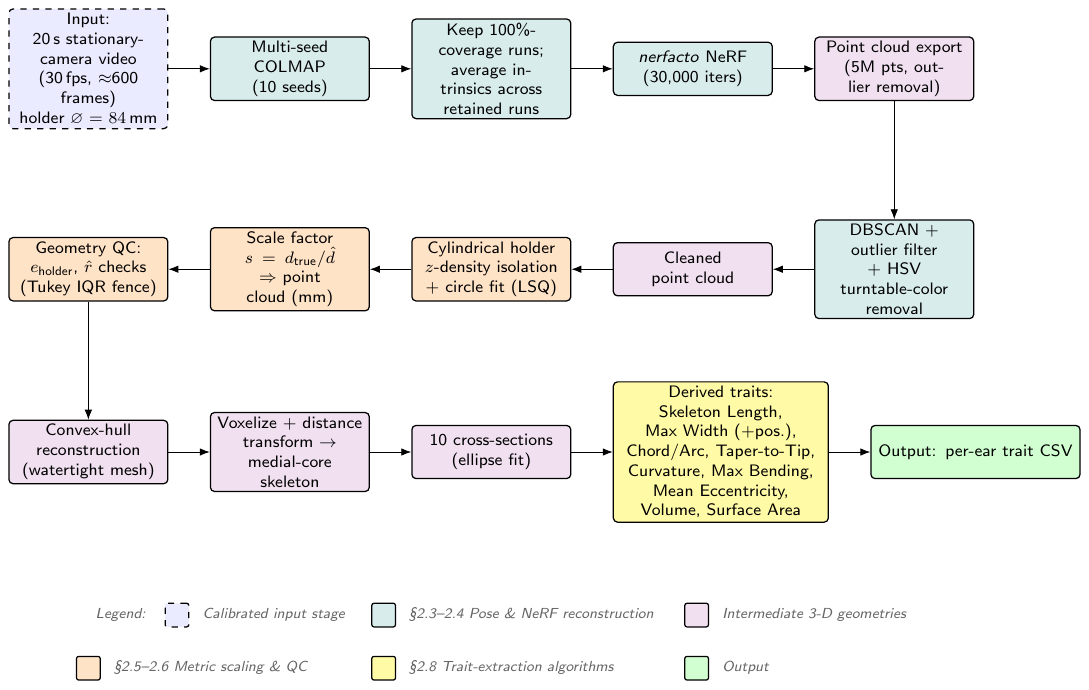}
\caption{\textbf{Algorithmic pipeline for 3-D maize ear phenotyping.} Camera poses from multi-seed COLMAP initialize \textit{nerfacto} NeRF training, which yields a dense point cloud. The point cloud is cleaned, isolated to the cylindrical holder for metric scaling, and passed through the automated geometry quality-control gate. The scaled point cloud is converted to a watertight convex-hull mesh, voxelized and skeletonized, and divided into ten cross-sections, from which the size and shape traits of \tabref{tab:trait_definitions} are computed and written to a per-ear CSV. }
    \label{fig:algo_diagram}
\end{figure*}

The overall workflow, from raw turntable video to a per-ear trait CSV, is summarized in \figref{fig:algo_diagram}. Camera poses are first estimated by the multi-seed COLMAP strategy (\secref{sec:colmapmethod}) and used to train a \textit{nerfacto} NeRF, from which a dense point cloud is exported. The point cloud is cleaned, metrically scaled using the cylindrical holder fiducial (\secref{sec:scaling}), and passed through an automated geometry quality-control gate (\secref{sec:qc}) before being converted to a watertight convex-hull mesh, skeletonized, and reduced to the trait set defined in \secref{sec:traits}. A complementary, qualitative view of the intermediate 3-D output at each of these stages is shown in \figref{fig:algo_setup} (\secref{sec:reconstruction}).

\subsection{Imaging hardware and cost breakdown}
\begin{figure}[t]
    \centering
    \includegraphics[width=1.0\textwidth]{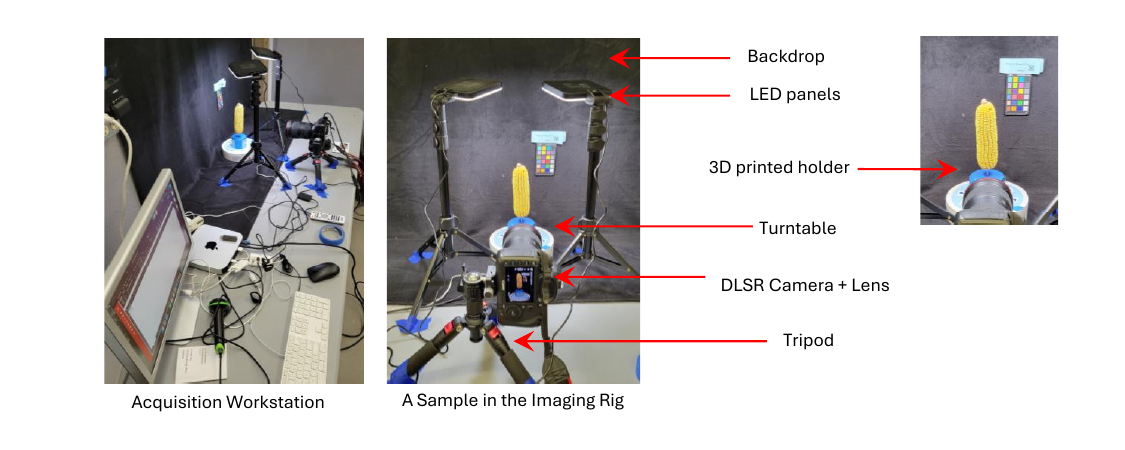}
\caption{\textbf{Imaging hardware for 3-D maize ear phenotyping.} The middle image illustrates the imaging setup: a fixed DSLR camera records a rotating maize ear mounted on a low-cost motorized turntable, illuminated by two off-axis LED panels against a matte-black backdrop. A color calibration chart ensures consistent photometric conditions across captures.}
    \label{fig:setup}
\end{figure}

\figref{fig:setup} illustrates the acquisition rig. A consumer-grade DSLR (Canon EOS 6D) fitted with a 50 mm focal-length lens is positioned a few decimeters from the specimen; the ear is mounted on a low-cost motorized turntable that rotates at approximately 3 rpm. The object is illuminated by two off-axis 12 V LED panels that provide uniform lighting, while a matte-black backdrop suppresses background clutter. A color-calibration chart is placed in the field of view to guarantee consistent photometric conditions across all captures. The sample holder is a cylindrical tube of 84\,mm (3.3\,in) outer diameter, 3-D printed in PLA. The cylindrical tube's diameter serves as a metric fiducial for later scaling. All imaging components are commercially available and together cost around 607 USD.

\begin{table}[t!]
\centering
\caption{Bill of Materials (USD).}
\label{tab:bill_of_materials}
\renewcommand{\arraystretch}{1.05}
\begin{tabular}{@{}llr@{}}
\toprule
\textbf{Component} & \textbf{Model / Specs}               & \textbf{Cost}  \\
\midrule
DSLR + lens                & Canon EOS 6D + 50 mm               & \$500  \\
Turn-table                 & PC-controlled turntable, 12.6 in platform, 350 lb load           & \$50   \\
LED panels (2)             & 12 V, 500 lm each                 & \$40   \\
Black backdrop             & Matte board, $1\,\mathrm{m}^{2}$   & \$10   \\
Cylindrical holder         & 3-D printed PLA, 3.3 in diameter  & \$7   \\
\midrule
\textbf{Total}             &                                    & \textbf{\$607}  \\
\bottomrule
\end{tabular}
\end{table}

\subsection{Data acquisition}
Maize ears were sourced from a diversity panel of 370 inbred maize genotypes (the SAM panel, Iowa State University), from which a random subset of 300 ears was selected for processing. Of the subset, 250 ears produced a complete end-to-end reconstruction; the remaining 50 ears were excluded automatically (15 at the COLMAP registration stage and 35 at the geometry quality-control stage) as described in \secref{sec:colmapmethod} and \secref{sec:qc}. Data collection spanned multiple sessions on different days and was carried out by 5 independent operators; five operators followed a fixed capture protocol across sessions. Two motorized turntable models from the same manufacturer were used across sessions; both operated at approximately 3\,rpm and produced comparable frame sequences.

For each ear, the holder was placed on the turntable, and the camera and turntable (both programmatically controlled from and connected to a PC) were triggered together for synchronized video capture over a full 360\textdegree{} rotation. Videos were captured at 30\,fps with a resolution of 1080p, yielding roughly 600 frames per ear (duration $\approx$20\,s). Each recording was automatically renamed with a unique per-ear identifier and uploaded to cloud storage for downstream reconstruction. The raw video files were subsequently processed with \texttt{ffmpeg} as follows: frames were extracted at 3\,fps, resulting in approximately 60 images per ear. The extracted image sequence together with the accompanying metadata form the input for the multi-pass COLMAP pose-estimation stage described in \secref{sec:colmapmethod}. All steps (synchronized capture, renaming, cloud upload, and frame extraction) are orchestrated by a small Python wrapper that creates the required directory hierarchy and logs the operations for reproducibility.

\begin{figure}[t]
    \centering
    \includegraphics[width=1.0\textwidth]{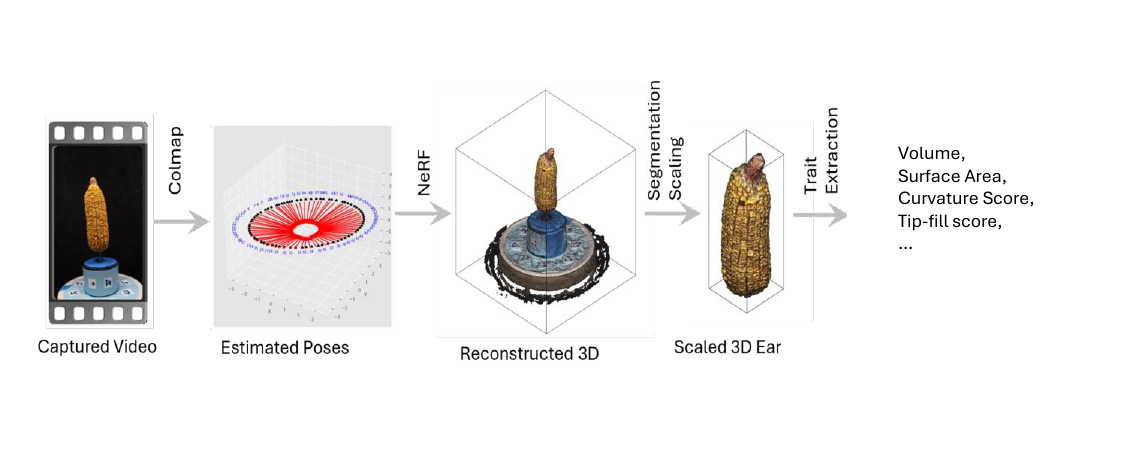}
\caption{\textbf{Visual walkthrough of the 3D maize ear reconstruction pipeline.} Whereas \figref{fig:algo_diagram} details the algorithmic steps of the pipeline, this figure shows the intermediate 3-D output produced at each stage. A 20-second video of the rotating ear is passed through COLMAP to estimate per-frame camera poses (Estimated Poses). The calibrated poses initialize a \textit{nerfacto} Neural Radiance Field, which produces a dense 3-D reconstruction of the ear and its cylindrical holder (Reconstructed 3-D). Segmentation isolates the ear from the holder and background, and the known holder diameter provides a metric scale factor (Scaled 3-D Ear). Morphological traits, including volume, surface area, curvature score, and additional shape descriptors, are then extracted from the scaled watertight mesh. }
    \label{fig:algo_setup}
\end{figure}

\subsection{3-D reconstruction}
\label{sec:reconstruction}
The 3-D reconstruction follows four sequential stages (\figref{fig:algo_setup}):
\begin{enumerate}
  \item Camera pose estimation from the video using COLMAP with
        multi-seed initialization.
  \item Volumetric reconstruction using a \textit{nerfacto} Neural
        Radiance Field trained on the estimated poses.
  \item Segmentation of the ear from the holder and background,
        followed by metric scaling using the known holder diameter.
  \item Trait extraction from the scaled watertight mesh.
\end{enumerate}

\subsubsection{Pose estimation: multi-seed COLMAP}
\label{sec:colmapmethod}
Camera poses were estimated using COLMAP \citep{Schoenberger2016_COLMAP} via the \texttt{ns-process-data} front-end. For each video, ten independent COLMAP reconstructions were run using a sequential matcher and the default SIFT \citep{Lowe2004_SIFT} ratio threshold (0.8); the only source of variability between runs was the random seed, which changes the order of feature-match sampling and RANSAC \citep{Fischler1981_RANSAC} initialization and can therefore produce different candidate models from the same input.

Post-processing proceeded as follows. For each run, frame-registration coverage was computed as
\[
\text{Coverage} = \frac{N_{\text{reg}}}{N_{\text{tot}}} \times 100\%.
\]
Runs with coverage below 100\,\% were discarded; if no run achieved full coverage, the ear was flagged for re-capture. Among retained runs, the run with the lowest mean reprojection error provided the camera extrinsics (per-frame pose transforms). Camera intrinsics (focal length, principal point, distortion coefficients) were averaged across all retained runs; in practice, intrinsic variation was small ($<$0.5\,\% relative change in focal length).

\subsection{Neural radiance field reconstruction}
The camera poses and intrinsics from \secref{sec:colmapmethod} were passed to \texttt{nerfstudio} (v0.4) using the \textit{nerfacto} model, which uses a multi-resolution hash-grid MLP to represent a continuous radiance field. Training ran for 30\,000 iterations (batch size 4096, learning rate $1\times10^{-2}$) on a single NVIDIA A100 (CUDA 12.1, PyTorch 2.2), requiring approximately fourteen minutes per ear.

The model optimizes a photometric loss with the inter-level and distortion regularizers introduced by \citet{Barron2022_MipNeRF360}:
\begin{equation}
  \mathcal{L} = \frac{1}{|R|}\sum_{\mathbf{r}\in R}
    \big\|\hat{C}(\mathbf{r}) - C_{\mathrm{gt}}(\mathbf{r})\big\|_2^{2}
    + \lambda_{\mathrm{inter}}\mathcal{L}_{\mathrm{inter}}
    + \lambda_{\mathrm{dist}}\mathcal{L}_{\mathrm{dist}},
\end{equation}
where $\hat{C}(\mathbf{r})$ is the rendered color for ray $\mathbf{r}$ and $C_{\mathrm{gt}}(\mathbf{r})$ is the observed pixel color.

After training, a dense point cloud (5M points, outlier removal enabled) is exported with \texttt{ns-export pointcloud} and passed to the segmentation and scaling stage.

\begin{figure*}[t]
    \centering
    \includegraphics[height=4.6cm]{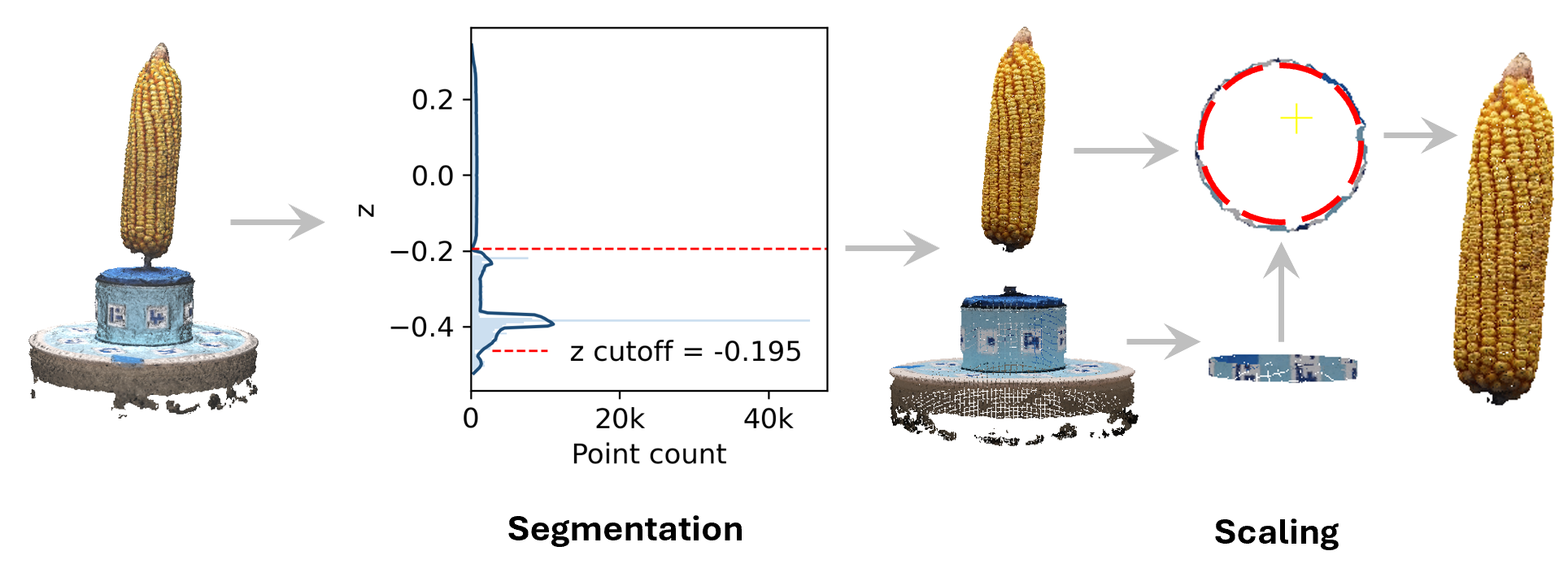}
\caption{\textbf{Automated metric scaling via the cylindrical holder.} \textit{Left:} Raw exported point cloud of the ear and its cylindrical holder. \textit{Middle:} Smoothed point-density profile along the vertical ($z$) axis; the red dashed line marks the automatically detected cutoff separating the ear from the holder base. \textit{Right:} Points in the stable holder region are isolated and fitted with a circle by algebraic least squares; the fitted diameter relative to the known physical diameter (84\,mm) yields a global scale factor applied to the entire reconstruction.}
    \label{fig:scaling_pipeline}
\end{figure*}

\subsection{Metric scaling using the cylindrical holder}
\label{sec:scaling}

The \textit{nerfacto} model reconstructs the scene in a normalized coordinate system; the exported point cloud therefore carries no intrinsic physical unit. We recover a metric scale factor automatically from the cylindrical sample holder, which is visible in every frame and whose outer diameter is known precisely ($d_{\text{true}} = 84$\,mm), eliminating the need for a separate calibration target. Systematic geometric distortion is mitigated upstream by the averaged camera intrinsics from the multi-seed COLMAP strategy (\secref{sec:colmapmethod}).

The cylindrical holder points were isolated by retaining those whose $z$-coordinates fall within the region of nearly constant point density along the vertical axis (\figref{fig:scaling_pipeline}, Middle). A thin horizontal slice $S = \{\mathbf{p} \mid |z - z_c| \le \delta\}$ centered at $z_c$ is projected onto the $xy$-plane and fitted with a circle by algebraic least squares \citep{Kanatani1998-af}, yielding estimated radius $r$ and diameter $\hat{d} = 2r$. The global scale factor is then
\begin{equation}
  s \;=\; \frac{d_{\text{true}}}{\hat{d}},
  \label{eq:scale}
\end{equation}
and multiplying all point-cloud coordinates uniformly by $s$ converts the reconstruction from normalized units to physical millimeters.

\begin{figure*}[t]
    \centering
    \includegraphics[width=1.0\textwidth]{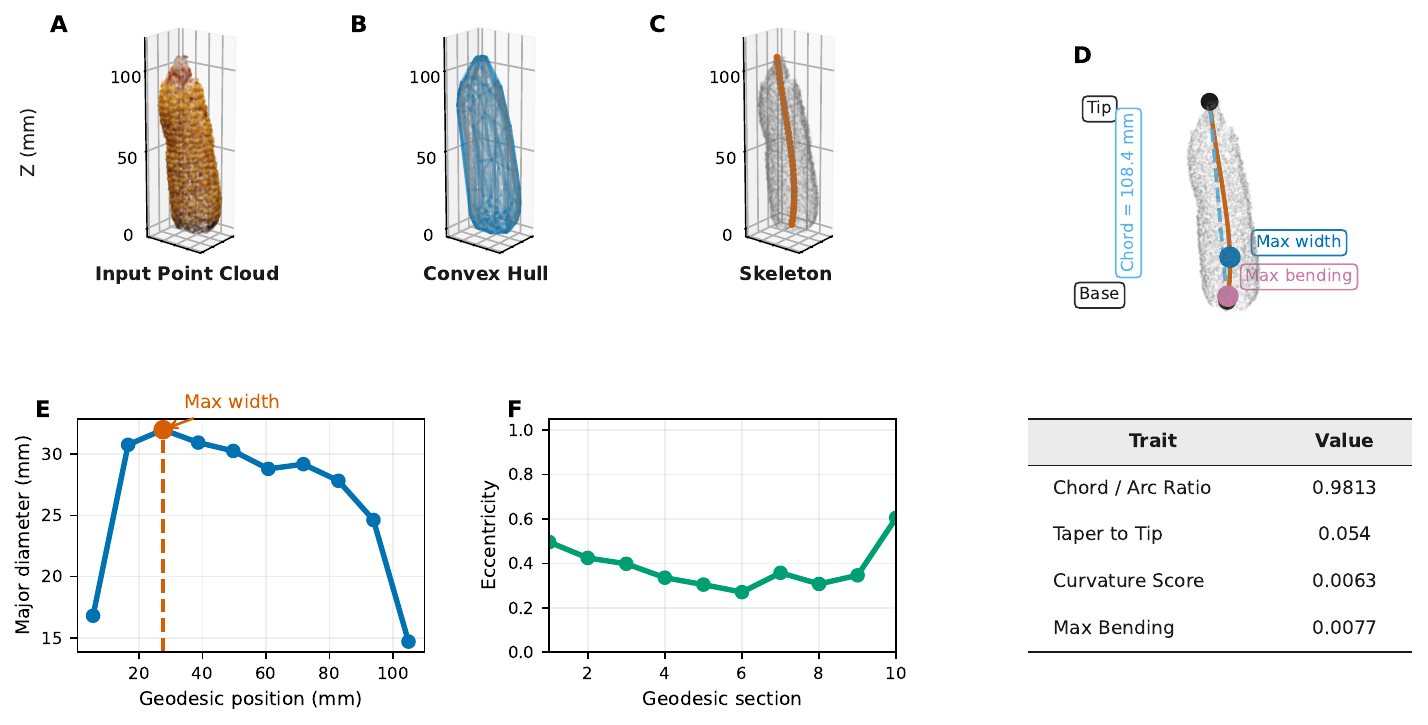}

\caption{\textbf{Three-dimensional maize ear trait extraction pipeline.} (A) Raw input point cloud with bounding box overlay. (B) Watertight solid mesh reconstructed from the point cloud; the mesh is converted to a volumetric solid prior to skeletonization to ensure topological correctness. (C) Topological skeleton extracted along the geodesic axis of the solid mesh, following the medial path from base to tip. (D) Trait landmark visualization showing key geometric measurements along the skeleton: the arc tracing the geodesic axis (red), the straight chord connecting tip to base (dashed blue), the point of maximum width (blue), and the point of maximum local bending (purple). (E) Major diameter profile measured at ten cross-sections distributed along the geodesic axis, with the position of maximum width indicated (dashed orange). (F) Cross-sectional eccentricity profile across the ten geodesic sections. Computed traits are summarized in the inset table: \textbf{Chord/Arc Ratio}, the ratio of the straight tip-to-base chord to the geodesic arc length ($= 1$ for a perfectly straight ear); \textbf{Taper to Tip}, the cross-sectional area at the distal section relative to the maximum section area (lower values indicate greater tapering); \textbf{Curvature Score}, the total turning angle along the skeleton normalized by geodesic length (higher values indicate more curved ears); and \textbf{Max Bending}, the strongest local bend rate at any single skeleton segment (higher values indicate a sharper localized bend).}
    \label{fig: trait extraction}
\end{figure*}

\subsection{Geometry quality control via cylindrical holder cross-section analysis}
\label{sec:qc}

Scale accuracy depends entirely on isolating the holder correctly, so the pipeline verifies that isolation before computing any trait. When the density profile along the vertical axis lacks a clear inflection point (due to partial holder occlusion, an ear positioned too close to the holder rim, or an unusually dense base region), the cutoff boundary can fall in the wrong location. This produces three classes of failure: the isolated region may capture only a partial ear, include structural parts of the ear body alongside the holder, or correspond entirely to the wrong region of the point cloud. In all cases, the circle fit is applied to an incorrect point set, yielding a holder cross-section that is non-circular, an erroneous scale factor, or both.

Two scalar indicators are computed from the fitted holder cross-section to detect segmentation failures automatically. The first is the holder eccentricity
\begin{equation}
  e_{\text{holder}} = \sqrt{1 - \left(\frac{\hat{b}}{\hat{a}}\right)^{2}},
  \label{eq:eholder}
\end{equation}
obtained by fitting a general conic shape \citep{Kanatani1998-af} to the projected holder slice, yielding semi-axes $\hat{a} \ge \hat{b} > 0$. A value of zero indicates a circular cross-section consistent with the physical holder; elevated values indicate that a non-holder region has been included in the fit. The second indicator is the fitted radius $\hat{r} = \hat{d}/2$ in NeRF normalized units. Because the camera, lens, and camera-to-object distance are held constant across sessions, $\hat{r}$ should fall within a narrow and reproducible range; a value that departs substantially from the expected range indicates that the circle fit has latched onto an entirely wrong region of the point cloud.

An ear is automatically flagged if $e_{\text{holder}}$ exceeds the upper Tukey IQR fence
\begin{equation}
  e_{\text{holder},\,i} > Q_3(e_{\text{holder}}) +
  1.5\;\mathrm{IQR}(e_{\text{holder}}),
  \label{eq:iqr}
\end{equation}
or if $\hat{r}$ falls outside the expected session range, evaluated across all ears that passed COLMAP registration within the same processing batch. Ears not flagged automatically, but whose holder cross-section appeared visually anomalous during routine inspection were also excluded and recorded in the rejection log. All excluded ears are written to a rejection log with their $e_{\text{holder}}$ and $\hat{r}$ values and the reason for exclusion. Representative failure cases are shown in the Supplementary \figref{fig:qc_failures}.

\begin{table}[t]
\centering
\caption{\textbf{Morphological trait definitions.} Cross-sections are indexed 1 (base) to 10 (tip) along the centreline skeleton. Size traits are reported in physical units; shape traits are dimensionless ratios unless stated otherwise.}
\label{tab:trait_definitions}
\setlength{\tabcolsep}{6pt}
\begin{tabular}{@{}l l p{6.8cm}@{}}
\toprule
\textbf{Trait} & \textbf{Unit} & \textbf{Definition} \\
\midrule
\multicolumn{3}{@{}l}{\textit{Size traits}} \\[2pt]
Skeleton Length        & mm            & Arc-length of the centreline skeleton from base to tip \\
Max Width              & mm            & Largest major diameter across all 10 cross-sectional ellipses \\
Max Width Position     & 0--1          & Fractional position of maximum width along the skeleton (0\,=\,base, 1\,=\,tip) \\
Convex-Hull Surface Area & cm$^{2}$    & Total surface area of the convex-hull mesh \\
Convex-Hull Volume     & cm$^{3}$      & Volume enclosed by the convex-hull mesh \\
BBox Height            & mm            & Height of the axis-aligned bounding box \\[6pt]
\multicolumn{3}{@{}l}{\textit{Shape traits}} \\[2pt]
Chord/Arc Ratio        & 0--1          & Straight tip-to-base chord divided by skeleton arc-length; $= 1$ for a perfectly straight ear \\
Taper to Tip           & 0--1          & Cross-sectional area at section~10 divided by the maximum section area; lower values indicate a sharper tip \\
Curvature Score        & rad\,mm$^{-1}$& Sum of skeleton turning angles divided by total arc-length; higher values indicate a more curved ear \\
Max Bending            & rad\,mm$^{-1}$& Peak local turning angle divided by local segment length; captures the sharpest single bend \\
Mean Eccentricity      & 0--1          & Mean $\sqrt{1-(b/a)^{2}}$ over 10 cross-sections; 0\,=\,circular, 1\,=\,flat ellipse \\[6pt]
\multicolumn{3}{@{}l}{\textit{Per-section traits} ($i = 1 \ldots 10$)} \\[2pt]
Section $i$ major axis & mm            & Major axis of the fitted ellipse at cross-section $i$ \\
Section $i$ minor axis & mm            & Minor axis of the fitted ellipse at cross-section $i$ \\
Section $i$ eccentricity & 0--1        & Eccentricity of the fitted ellipse at cross-section $i$ \\
\bottomrule
\end{tabular}
\end{table}

\subsection{Trait extraction}
\label{sec:traits}
Following metric scaling, the point cloud is processed through a series of geometric operations to extract morphological traits that characterize both the size and shape of the ear (\tabref{tab:trait_definitions}). Trait definitions follow the vocabulary established by EarCV \citep{Gonzalez2022-fd} and EARBOX \citep{Oury2021-zt}. The pipeline reports eleven whole-ear traits, six describing size and five describing shape, together with per-section values at ten cross-sections. The six size traits (skeleton length, maximum width, maximum-width position, convex-hull surface area, convex-hull volume, and bounding-box height) describe the overall dimensions of the ear. Shape traits capture geometry independently of absolute size: the Chord/Arc Ratio quantifies straightness; Taper to Tip describes the taper profile toward the tip; the Curvature Score and Max Bending measure axial bending; and Mean Eccentricity characterizes the cross-sectional roundness at ten sections distributed from base to tip. Each processing step is illustrated in \figref{fig: trait extraction}.

\paragraph{Point-cloud cleaning (Panel~A).}
Before any geometric analysis, the raw point cloud is subjected to three automatic cleaning steps. First, DBSCAN clustering \citep{Ester1996_DBSCAN} ($\varepsilon = 5\;\text{mm}$, minimum cluster size~50) isolates the largest connected component, discarding stray fragments and calibration artifacts. Second, a statistical outlier filter removes points whose mean distance to their $k = 20$ nearest neighbors exceeds $2\,\sigma$ above the population mean, suppressing reconstruction noise on the surface. Third, turntable reflections are eliminated by a two-pass HSV color filter: the first pass removes globally blue pixels (hue $180$--$270^{\circ}$, saturation $> 0.15$, value $< 0.85$), and a second pass applies a relaxed saturation threshold (saturation $> 0.08$) to the bottom~25\% of the scene where residual turntable color is most prevalent. The cleaned point cloud (Panel~A) is carried forward to all subsequent steps.

\paragraph{Convex-hull reconstruction (Panel~B).}
The cleaned point cloud is converted directly to a three-dimensional convex hull using the incremental convex-hull algorithm in Open3D \citep{Zhou2018_Open3D}. The resulting mesh $\mathcal{M}=(\mathcal{V},\mathcal{F})$ is watertight by construction, requires no normal estimation or surface-fitting parameters, and encloses the ear tightly while remaining robust to surface irregularities.

\paragraph{Solid interior and skeleton extraction (Panel~C).}
The convex-hull mesh is voxelized on a regular grid with spacing $\Delta=3\;\text{mm}$. For each grid point, occupancy is determined by ray-casting (threshold $>0.5$), yielding a binary solid array $\mathcal{S}\subset\mathbb{Z}^{3}$. A Euclidean distance transform~\citep{Maurer2003-vw} is computed over $\mathcal{S}$, giving the distance from every interior voxel to the nearest surface. For each slice along the longitudinal axis (identified as the axis of greatest bounding-box extent), voxels whose distance-transform value exceeds $0.75$ of the slice maximum are retained as the \emph{medial core}, and their centroid is computed with weights proportional to the squared distance value. The resulting centerline is smoothed by fitting a cubic polynomial in each transverse dimension as a function of the longitudinal coordinate, and near-duplicate consecutive points (separation $< 0.25\,\Delta$) are removed. The arc length of the smoothed centerline defines the skeleton length~$L_{\text{skel}}$.

\paragraph{Global geometry (Panel~D).}
The mesh is re-oriented so that the longitudinal axis coincides with $z$ via principal component analysis. The skeleton arc length $L_{\text{skel}}$ serves as the primary measure of ear length. The straight-line distance between the skeleton endpoints defines the chord length $L_{\text{chord}}$. Surface area,~$A$, is the sum of triangle areas over all faces of the convex hull. Volume,~$V$, is obtained from the signed tetrahedral decomposition
\begin{equation}
  V = \frac{1}{6}\!\sum_{(i,j,k)\in\mathcal{F}}
      \det\!\bigl[\mathbf{v}_i,\mathbf{v}_j,\mathbf{v}_k\bigr],
\end{equation}
applied to the convex-hull mesh, which is guaranteed watertight. The maximum major diameter across all ten cross-sections (\secref{par:cross}) defines the \emph{max width}, and its relative position along the skeleton is also recorded.

\paragraph{Cross-sectional shape (Panels~E--F).}
\label{par:cross}
The skeleton is divided into $K = 10$ equal-length sections. At each section~$t$, mesh vertices within $\max(2.5\%\times L_{\text{skel}},\,2\;\text{mm})$ of the section plane are projected onto the plane perpendicular to the local skeleton tangent. A two-dimensional convex hull is computed from the projected points, and a direct least-squares ellipse is fitted to its boundary vertices \citep{Fitzgibbon1999-xd}. If the algebraic fit fails, a covariance-based fallback estimates semi-axes as $a_t = 2\sqrt{\lambda_1^{(t)}}$ and $b_t = 2\sqrt{\lambda_2^{(t)}}$ from the eigenvalues $\lambda_1^{(t)} \ge \lambda_2^{(t)}$ of the $2\times2$ scatter matrix.

Three section-wise quantities are reported: the major and minor diameters $D_{\text{maj}}^{(t)} = 2a_t$ and $D_{\text{min}}^{(t)} = 2b_t$; the cross-sectional eccentricity
\begin{equation}
  e_t = \sqrt{1 - \left(\frac{b_t}{a_t}\right)^{2}},
\end{equation}
which equals zero for a circular section and approaches unity for a highly elongated one; and the cross-sectional area obtained from the two-dimensional convex hull.

\paragraph{Derived shape descriptors.}
We use five shape descriptors to summarize ear morphology at the whole-ear level. The Chord/Arc Ratio $L_{\text{chord}} / L_{\text{skel}}$ quantifies straightness and equals unity for a perfectly straight ear.
The Taper to Tip index
\begin{equation}
  \text{Tap} = \frac{A_{K}}{\max_t A_t}
\end{equation}
expresses the cross-sectional area at the distal (tip) section relative to the maximum section area, with lower values indicating greater tapering. The Curvature Score
\begin{equation}
  \text{Cur} = \frac{\displaystyle\sum_{t=1}^{K-1} \Delta\theta_t}
                    {L_{\text{skel}}}
\end{equation}
accumulates the turning angle $\Delta\theta_t$ between consecutive skeleton tangent vectors and normalises by arc length; higher values indicate a more curved ear. The Max Bending is the peak local curvature along the skeleton,
\begin{equation}
  \kappa_{\max} = \max_t \frac{\Delta\theta_t}{\Delta s_t},
\end{equation}
where $\Delta s_t$ is the segment length between consecutive skeleton points; it captures the single sharpest bend. Finally, Mean Eccentricity $\bar{e} = K^{-1}\sum_{t=1}^{K} e_t$ summarizes the average departure from circularity across all ten sections.

All traits (global summaries and per-section values) are written to a single CSV file per ear. The full extraction pipeline is wrapped in a single command. Most trait vocabulary is inspired by the conceptual framework established by EarCV \citep{Gonzalez2022-fd} and EARBOX \citep{Oury2021-zt}; the principal contribution lies in the integration, automation, and reproducibility packaging. Two traits warrant specific framing. First, the Chord/Arc Ratio is a formally defined 3-D shape descriptor with no direct analogue in prior ear-phenotyping tools: 2-D imaging records only the straight-line projection of the ear axis, making the geodesic arc (and therefore any arc-to-chord comparison) inaccessible; the metric is the reciprocal of the tortuosity index used in vascular biology and movement ecology \citep{Hart1999_Tortuosity,Benhamou2004_Straightness}, here applied for the first time to quantify whole-ear axial straightness from a watertight mesh skeleton. Second, \textbf{Taper-to-Tip} is retained as a geometric descriptor of distal ear cross-sectional area but is explicitly interpreted as a shape proxy rather than a direct measure of kernel fill: without kernel-level segmentation of the mesh, it cannot distinguish natural geometric taper from tip kernel abortion, unlike the pixel-based fill scores of 2-D tools such as EarCV.

\begin{figure*}[tb]
    \centering
    \includegraphics[width=0.7\textwidth]{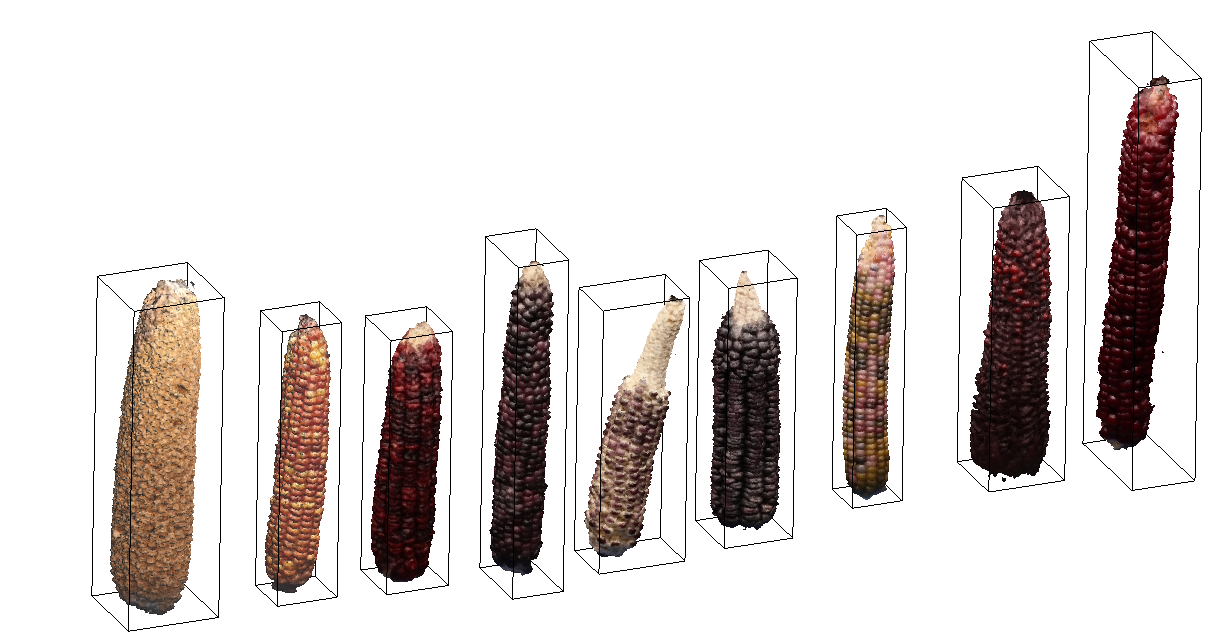}
\caption{ Example 3-D reconstructions of maize ears generated by the proposed workflow. The method robustly recovers ear geometry across substantial variations in size, shape, curvature, and kernel color. Each ear is shown together with its automatically estimated bounding box, which defines the spatial extent used for subsequent trait extraction and comparative analysis. The diversity of reconstructed ears illustrates that the pipeline generalizes well to heterogeneous phenotypes and captures both morphological and color variation in a consistent 3-D representation. }
    \label{fig:maize_ear_reconstructions}
\end{figure*}

\section{Results}
\label{sec:results}

\subsection{COLMAP registration and geometry quality control}
\label{sec:results_colmap}

\begin{figure*}[tp]
    \centering
    \includegraphics[width=1.0\textwidth]
{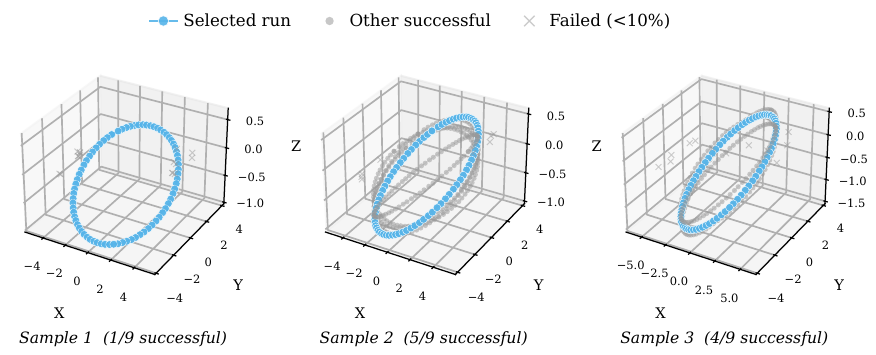}

\caption{\textbf{Run-to-run variability in camera pose trajectories} (A) 100\% seed convergence to consistent trajectories (B) Partial registration on some seeds (C) Inconsistent trajectories; ear flagged for re-capture }
    \label{fig:colmap poses}
\end{figure*}

\tabref{tab:colmap_success} summarizes the pose-estimation of 300 processed ears. Of the ten independent COLMAP runs launched per ear, at least one achieved 100\,\% frame registration for 285 ears (95\,\%); the remaining 15 ears (5\,\%) required re-capture because none of the ten runs produced a fully registered reconstruction. \figref{fig:colmap poses} illustrates the run-to-run variability in camera pose trajectories across three representative ears. The three examples span the range of outcomes: a case where all ten seeds converge to consistent trajectories, a case with partial registration on some seeds, and a case with inconsistent trajectories that caused the ear to be flagged for re-capture. The multi-seed strategy is effective precisely because individual seeds can fail on challenging ears while other seeds succeed; retaining only 100\,\%-coverage reconstructions and selecting the lowest-reprojection-error run provides a stable pose set for NeRF training.
\begin{table}[t!]
  \centering
\caption{COLMAP success statistics (multi-seed initialization).}
  \label{tab:colmap_success}
  \begin{tabular}{@{}l r r@{}}
    \toprule
    \textbf{Number of videos} & \textbf{100\,\% registration} & \textbf{Re-capture needed} \\
    \midrule
    300 & 285 (95\,\%) & 15 (5\,\%) \\
    \bottomrule
  \end{tabular}
\end{table}

After registration, the 285 ears were passed through the automated geometry quality-control step (\secref{sec:qc}).

In total, 35 ears were excluded at this stage. After applying the geometry quality-control filter, 250\,/\,300 ears (83.3\,\%) passed to trait extraction. A set of excluded cases are shown in the Supplementary \figref{fig:qc_failures}.

\begin{figure*}[tp]
    \centering

     \includegraphics[width=1.0\textwidth]{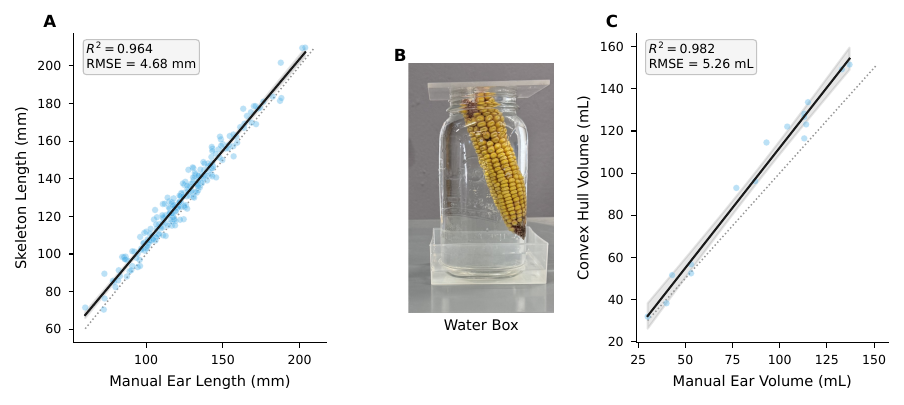}

\caption{
\textbf{(A)}~Automated skeleton length plotted against manual caliper measurement for all 250 ears ($R^{2} = 0.964$, $\mathrm{RMSE} = 4.68$\,mm). Skeleton length traces the geodesic arc along the curved ear axis, which exceeds the straight-line chord measured by calipers for bent ears; the systematic divergence with curvature is quantified in \figref{fig:traits_vs_length_error}.
\textbf{(B)}~Photograph of the water-displacement apparatus. Each ear was submerged in a container filled to the brim and held flush with the opening by an acrylic plate; displaced water was collected and read to the nearest 1\,mL in a graduated 100\,mL cylinder.
\textbf{(C)}~Convex-hull volume derived from the reconstructed point cloud plotted against manually measured displacement volume ($R^{2} = 0.982$, $\mathrm{RMSE} = 5.26$\,mL). The regression line lies slightly above the 1:1 reference (dotted), reflecting a small positive bias attributable to the convex hull enclosing empty space within the concavity of curved ears.
In panels~(A) and (C), the solid line is the ordinary least-squares fit, the shaded band the 95\,\% confidence interval of the mean, and the dotted line the 1:1 reference; $R^{2}$ and RMSE are inset. }

    \label{fig: ground truth}
\end{figure*}

\subsection{Diverse Maize Ears (Qualitative Examples)}
\figref{fig:maize_ear_reconstructions} shows 3-D reconstructions of a curated set of diverse maize ears (different from the 300 ears used for quantitative experiments). The pipeline recovers ear geometry across substantial variation in size, shape, curvature, and kernel color. Each ear is shown with its automatically estimated axis-aligned bounding box. The reconstructions illustrate that the pipeline generalizes to heterogeneous phenotypes without manual parameter tuning: short and long ears, straight and strongly curved ears, and ears ranging from yellow to red-purple kernel pigmentation are all successfully reconstructed. These qualitative results support the quantitative validation in the following section.

\begin{figure*}[t!]
\centering
\includegraphics[width=0.8\textwidth]{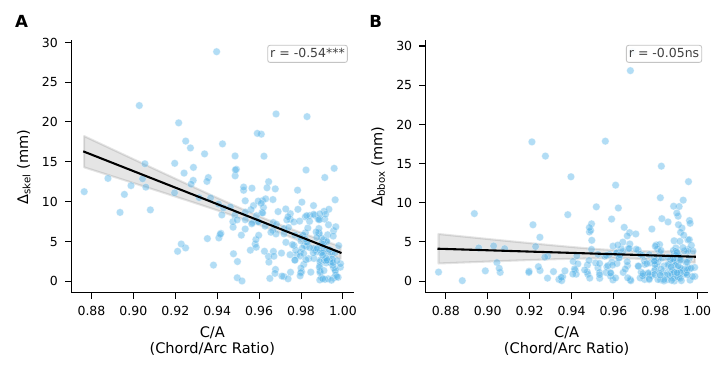}
\caption{
\textbf{(A)}~Absolute deviation of skeleton length from manual caliper measurement ($\Delta_{\text{skel}}$, mm) plotted against Chord/Arc ratio (C/A) for all 250 ears ($r = -0.54$, $p < 0.001$). Ears with lower C/A (indicating greater longitudinal curvature) show systematically larger disagreement between skeleton length and caliper measurement.
\textbf{(B)}~Absolute deviation of bounding-box height from manual caliper measurement ($\Delta_{\text{bbox}}$, mm) plotted against the same C/A axis ($r = -0.05$, ns). The regression slope is near zero, confirming that bounding-box height error is insensitive to ear curvature. The contrasting responses in panels~(A) and (B) establish that the caliper--skeleton discrepancy arises from a difference in measurement definition: bounding-box height and manual calipers both record the straight-line chord from base to tip, whereas skeleton length traces the geodesic arc, which necessarily exceeds the chord as curvature increases.
Significance codes: $^{***}p < 0.001$; ns~$p \ge 0.05$. $\Delta_{\text{skel}} = |\text{skeleton length} - \text{manual}|$; $\Delta_{\text{bbox}} = |\text{bounding-box height} - \text{manual}|$. }
\label{fig:traits_vs_length_error}
\end{figure*}

\subsection{Geometric validation}
\label{sec:vol_validation}

\figref{fig: ground truth} summarizes geometric validation of the pipeline against two independent manual references.

\paragraph{Skeleton length}
Automated skeleton length was compared against manual caliper measurements for all $n = 250$ ears (\figref{fig: ground truth}(A)), yielding $R^{2} = 0.964$ and $\mathrm{RMSE} = 4.68$\,mm, of the same order as the $\mathrm{RMSE} = 4.4$\,mm reported by \citet{Oury2021-zt}, although $R^{2}$ values are not directly comparable across panels with different trait ranges. The residual error reflects two contributions: natural caliper measurement variability and a fundamental difference in what each method measures. Manual calipers record the straight-line chord from base to tip; skeleton length traces the geodesic arc along the curved ear axis, which exceeds the chord for bent ears. \figref{fig:traits_vs_length_error} quantifies this distinction. Panel~(A) shows a strong negative relationship between $\Delta_{\text{skel}}$ and Chord/Arc ratio ($r = -0.54$, $p < 0.001$): ears with lower C/A (indicating greater curvature) exhibit systematically larger skeleton-length errors relative to the caliper reference. Panel~(B) shows that bounding-box height error ($\Delta_{\text{bbox}}$) is insensitive to C/A ($r = -0.05$, ns), with the regression slope near zero across the full curvature range. The divergent responses confirm that the caliper--skeleton discrepancy is a measurement-definition artifact rather than a reconstruction error: bounding-box height and manual calipers both measure the chord, so they agree regardless of ear shape, whereas skeleton length measures the arc and necessarily diverges from the chord as curvature increases.

\paragraph{Convex-hull volume}
A subset of $n = 15$ ears spanning the full size range was measured by water displacement. Each ear was submerged in a container filled to the brim and held flush with the opening by an acrylic plate (\figref{fig: ground truth}(B)); displaced water was collected and read to the nearest 1\,mL in a graduated 100\,mL cylinder. This method provides a direct physical volume reference independent of any geometric assumption. Convex-hull volume agrees closely with displacement volume (\figref{fig: ground truth}(C); $R^{2} = 0.982$, $\mathrm{RMSE} = 5.26$\,mL), though the regression line lies slightly above the 1:1 reference, reflecting a small positive bias: the convex hull encloses empty space within the concavity of curved ears, contributing to hull volume but not to displacement. This is an inherent limitation of the convex-hull estimator rather than a reconstruction error, consistent with the curvature-driven discrepancies previously identified for skeleton length measurements. Because this bias stems from the convexity of the hull itself rather than from imprecision in integrating its volume, eliminating it would require replacing the convex hull with a non-convex, concavity-preserving surface reconstruction, which is planned for future work.

\subsection{Population-level trait distributions}
\begin{figure*}[tb]
    \centering
    \includegraphics[width=\textwidth]{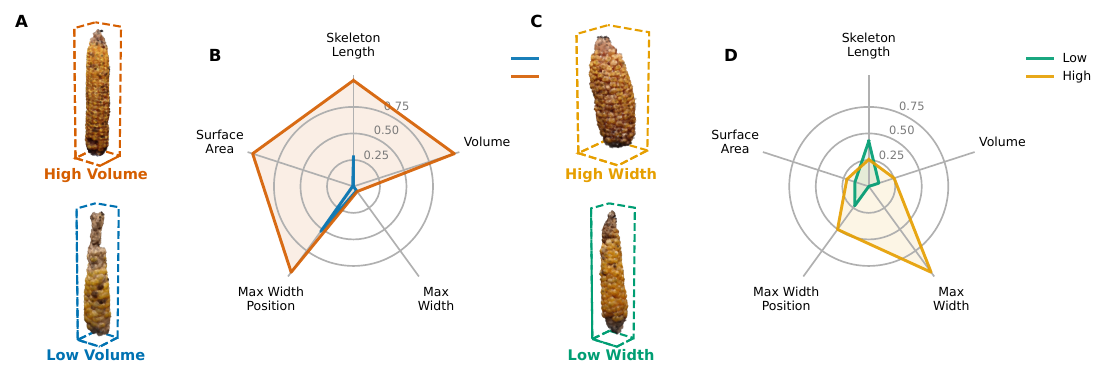}
    \caption{\textbf{Size-trait variation across the 3D maize ear population illustrated by contrasting ear pairs.} \textbf{(A)}~RGB photographs of a large, well-filled ear (orange bounding box) and a small, sparse ear (blue bounding box), selected to contrast in volume. \textbf{(B)}~Radar chart overlaying the normalised trait profiles of the two ears in panel~(A) across five size descriptors (Skeleton Length, Volume, Max Width, Max Width Position, Convex-Hull Surface Area; axes normalised to the population range). The large ear scores broadly across all size traits; the small ear is uniformly compact. \textbf{(C)}~RGB photographs of a wide ear (yellow bounding box) and a narrow ear (green bounding box), selected to contrast in maximum width. \textbf{(D)}~Radar chart for the ear pair in panel~(C), showing that maximum width co-varies with volume and surface area but not proportionally with skeleton length, indicating semi-independent variation in width and length across the panel. Ear photographs are not shown at a common scale; each image preserves the aspect ratio of the individual ear. }
    \label{fig:size_extremes}
\end{figure*}

\figsref{fig:size_extremes}{fig:shape_extremes} illustrate the phenotypic diversity captured by the pipeline through pairs of morphologically contrasting ears drawn from the size and shape trait spaces, respectively. Each pair is accompanied by a radar chart overlaying their normalized profiles across a subset of that group (five of the six size traits, four of the five shape traits; bounding-box height and Max Bending are omitted for legibility), revealing how variation in one trait co-varies with the remaining descriptors.

For size traits (\figref{fig:size_extremes}), a large, well-filled ear and a small, sparse ear (panels~A--B) differ markedly across skeleton length, volume, and surface area simultaneously, consistent with overall ear size being the dominant axis of co-variation within this group. A wide ear and a narrow ear (panels~C--D) show a more differentiated profile: the wide ear scores broadly across all size traits, while the narrow ear retains moderate skeleton length, indicating that width and length can vary semi-independently across the diversity panel.

\begin{figure*}[tb]
    \centering
    \includegraphics[width=\textwidth]{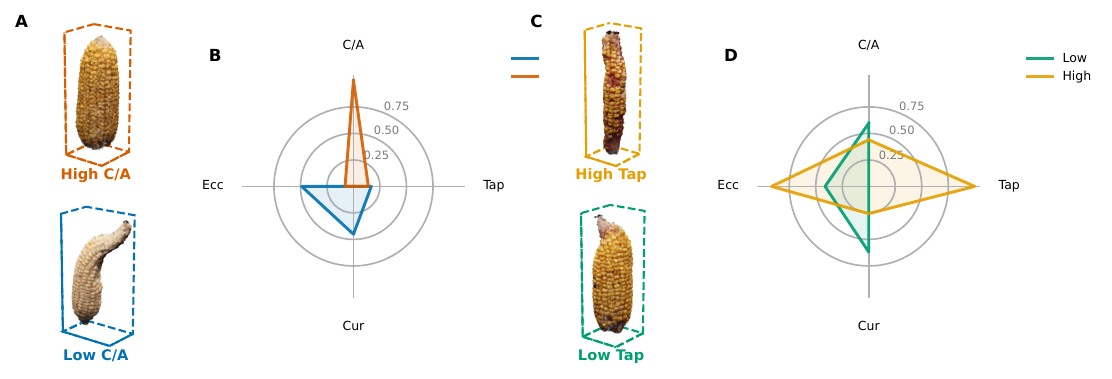}
    \caption{\textbf{Shape-trait variation across the 3D maize ear population illustrated by contrasting ear pairs.} \textbf{(A)}~RGB photographs of a straight ear (orange bounding box; C/A $\approx 1$) and a curved ear (blue bounding box), selected to contrast in Chord/Arc ratio. \textbf{(B)}~Radar chart overlaying the normalised shape profiles of the two ears in panel~(A) across four shape descriptors (C/A, Tap, Cur, Ecc; axes normalised to the population range). The two ears diverge on C/A while sharing comparable values for the remaining traits, confirming that curvature is an independent morphological dimension. \textbf{(C)}~RGB photographs of a cylindrical ear with sustained distal width (yellow bounding box) and a strongly tapered ear (green bounding box), selected to contrast in Taper-to-Tip. \textbf{(D)}~Radar chart for the ear pair in panel~(C), illustrating that Taper-to-Tip varies semi-independently of the other shape descriptors. Ear photographs are not shown at a common scale; each image preserves the aspect ratio of the individual ear. \textit{Trait abbreviations}: \textbf{C/A}: Chord/Arc Ratio ($=1$ for a perfectly straight ear); \textbf{Tap}: Taper-to-Tip (lower $=$ greater tapering); \textbf{Cur}: Curvature Score (higher $=$ more curved); \textbf{Ecc}: Mean Cross-sectional Eccentricity (higher $=$ more elliptic cross-section). }
    \label{fig:shape_extremes}
\end{figure*}

For shape traits (\figref{fig:shape_extremes}), a straight ear and a curved ear (panels~A--B) diverge primarily along the C/A axis while sharing comparable Tap and Ecc values, confirming that curvature is captured as an independent morphological dimension. A cylindrical ear with sustained distal width and a strongly tapered ear (panels~C--D) illustrate the full range of the Taper-to-Tip trait.

\begin{figure*}[tb]
    \centering
    \includegraphics[width=\textwidth]{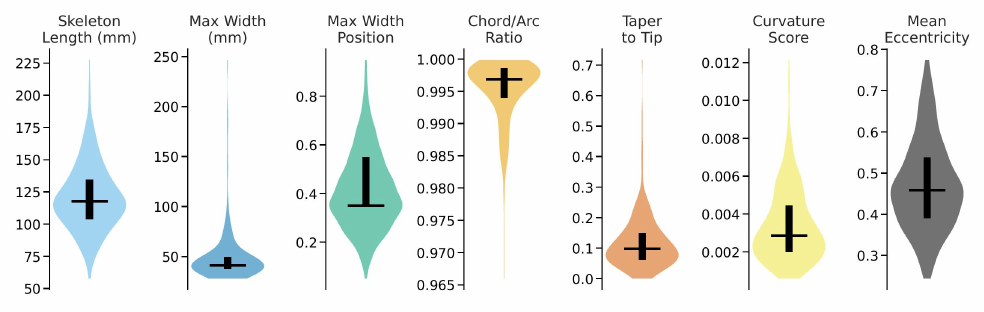}
    \caption{\textbf{Population-level distributions of seven of the eleven whole-ear morphometric traits ($n = 250$).} Violin plots for skeleton length, maximum width, maximum width position, Chord/Arc ratio, Taper-to-Tip, Curvature Score, and mean cross-sectional eccentricity. The horizontal bar and crosshair inside each violin mark the median and interquartile range, respectively. Chord/Arc ratio is tightly clustered near unity, indicating that most ears are nearly straight; Taper-to-Tip and Curvature Score are right-skewed, reflecting that strong tapering and high curvature are minority phenotypes within this panel. All remaining traits show broad, continuous variation, confirming that the pipeline discriminates meaningfully across the diversity panel. }
    \label{fig:trait_distributions}
\end{figure*}

\figref{fig:trait_distributions} shows the population-level distributions of seven of the eleven whole-ear traits as violin plots ($n = 250$); the remaining four are reported in the per-ear CSV output. All traits show substantial spread across the diversity panel. Skeleton length ranges from approximately 75 to 225\,mm with a roughly symmetric distribution, while maximum width is right-skewed, with most ears narrow and displaying a long upper tail. Chord/Arc ratio is tightly clustered near unity (0.965--1.000), indicating that the majority of ears are nearly straight, with curvature representing a minority phenotype. Taper-to-Tip and Curvature Score are each right-skewed, reflecting that strong tapering and high curvature are less common within this panel. Mean Eccentricity is broadly and approximately symmetrically distributed around 0.45, indicating that elliptic cross-sections are prevalent throughout the population.

\begin{figure*}[tb]
    \centering
    \includegraphics[width=\textwidth]{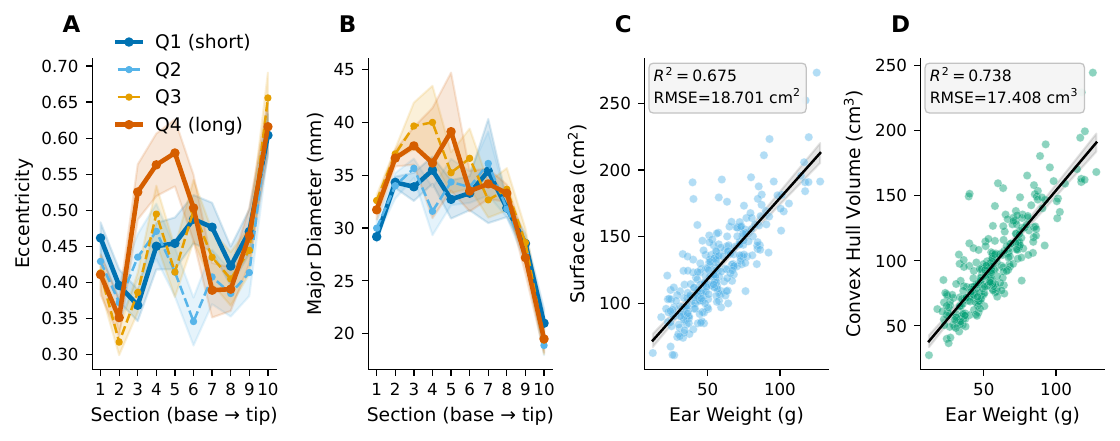}
\caption{\textbf{Cross-sectional profiles and biological associations of 3-D morphometric traits.} \textbf{(A)}~Mean cross-sectional eccentricity plotted from base to tip (sections 1--10) stratified by ear-length quartile (Q1: short; Q4: long). Profiles are consistent across quartiles, with a shared mid-ear peak near section~5 and convergence toward the tip, indicating that cross-sectional shape is conserved across ear sizes. \textbf{(B)}~Mean major-axis diameter (mm) along the same ten sections, stratified by quartile. Longer ears show larger diameters throughout, confirming that ear length and girth co-vary within this panel. \textbf{(C)}~Convex-hull surface area plotted against manually measured ear weight ($R^{2} = 0.675$, $\mathrm{RMSE} = 18.70$\,cm$^{2}$). \textbf{(D)}~Convex-hull volume plotted against ear weight ($R^{2} = 0.738$, $\mathrm{RMSE} = 17.41$\,cm$^{3}$). The moderate associations in panels~(C) and~(D) reflect a general size effect; residual variance is attributable to differences in kernel moisture and maturity state at the time of weighing rather than geometric inaccuracy. In all profile panels, shaded bands show the 95\,\% confidence interval of the mean. }
    \label{fig:trait_profiles}
\end{figure*}

\figref{fig:trait_profiles} provides further characterization of cross-sectional geometry and its association with ear size. Panels~(A) and~(B) show mean eccentricity and major-axis diameter profiles from base to tip, stratified by ear-length quartile. Eccentricity profiles are largely consistent across quartiles, with a shared mid-ear peak near section~5 and convergence toward the tip, suggesting that cross-sectional shape is conserved regardless of overall ear length. Major-axis diameter profiles show a clear quartile ordering throughout, confirming that longer ears are proportionally wider. Panels~(C) and~(D) show the association between convex-hull surface area and convex-hull volume against manually measured ear weight, yielding $R^{2} = 0.675$ ($\mathrm{RMSE} = 18.70$\,cm$^{2}$) and $R^{2} = 0.738$ ($\mathrm{RMSE} = 17.41$\,cm$^{3}$), respectively. The moderate associations reflect a general size effect, with residual variance attributable to differences in kernel moisture and maturity state at the time of weighing rather than geometric inaccuracy.

\subsection{Human-effort comparison}
\tabref{tab:human_effort} contrasts operator time for the conventional manual workflow with the proposed automated pipeline. Manual phenotyping (comprising caliper measurements of ear length, visual scoring, and data entry) requires approximately five minutes per ear. The automated pipeline reduces hands-on involvement to approximately one minute per ear (mounting the sample and initiating video capture); all downstream steps, including structure-from-motion reconstruction, point-cloud extraction, meshing, skeletonization, and trait computation, run unattended on a GPU node. Crucially, the pipeline simultaneously delivers traits (skeleton length, convex-hull volume, surface area, cross-sectional profiles, and five shape descriptors) that are inaccessible to manual caliper measurement.

\begin{table}[t!]
  \centering
\caption{Human-effort comparison (per ear).}
  \label{tab:human_effort}
  \begin{tabular}{@{}l r r@{}}
    \toprule
    \textbf{Task} & \textbf{Manual workflow} & \textbf{Proposed pipeline} \\
    \midrule
    Total operator time & $\approx$5 min & $\approx$1 min \\
    \bottomrule
  \end{tabular}
\end{table}

Taken together, the validation results, population-level trait distributions, and biological associations confirm that the low-cost, fully automated pipeline delivers geometrically accurate and biologically relevant 3-D phenotypes across a diverse maize panel, while reducing operator involvement by approximately 80\,\% relative to conventional manual measurements.

\section{Discussion}
\label{sec:discussion}

\subsection{Practical impact}
The system presented here meets three practical requirements for deployment in a breeding program: (i)~a low-cost hardware platform (\$607 total; \tabref{tab:bill_of_materials}), (ii)~an approximately 80\,\% reduction in operator time relative to manual caliper measurement (\tabref{tab:human_effort}), and (iii)~measurements validated against independent manual references (\secref{sec:vol_validation}). The imaging rig comprises a commodity DSLR, a motorized turntable, two LED panels, and a 3D-printed cylindrical sample holder, all components obtainable from standard laboratory suppliers. Following a 20-second video capture, the workflow executes autonomously on a GPU node, freeing the operator to prepare the next sample. The pipeline may be applicable to medium-scale breeding programs that require phenotypic characterization of hundreds of ears per season, though throughput at larger scales and performance across diverse environments remain to be evaluated.

\subsection{Design advantages}
Several technical choices contribute to the reliability of the approach.

\paragraph{Fixed-camera, rotating object.}
A stationary camera reduces the need for multi-viewpoint synchronization and simplifies mechanical calibration. Constraining object motion to a single rotation axis limits the camera pose search space, which may improve reconstruction consistency across captures, though this has not been formally compared against a moving-camera setup.

\paragraph{Multi-seed COLMAP robustness.}
Running COLMAP ten times with different random seeds (\secref{sec:colmapmethod}) and retaining only reconstructions with 100\,\% frame registration provides a safeguard against occasional feature-matching failures caused by low-texture or specular kernels. Averaging intrinsic parameters across successful runs reduces sensitivity to any single reconstruction and provides a more stable initialisation for subsequent NeRF training.

\paragraph{What the 3D mesh enables.}
Bounding-box height is computable from any point cloud and agrees closely with manual caliper measurements because both measure the straight-line chord from base to tip. Skeleton length, by contrast, traces the geodesic arc along the curved ear axis and diverges from the caliper reference predictably with curvature (\figref{fig:traits_vs_length_error}); computing it requires skeletonization of a closed surface. The watertight mesh further enables (i)~volume estimation via the signed tetrahedral formula, (ii)~total surface area from triangle sums, and (iii)~cross-sectional shape profiles that are less sensitive to the surface holes that can affect point-cloud slicing. These traits are not straightforwardly recoverable from a single 2D image, though direct comparison with alternative 3D acquisition methods is outside the scope of this work. Validation against water-displacement volume yields $R^{2} = 0.982$ and $\mathrm{RMSE} = 5.26$\,mL (\figref{fig: ground truth}(C)), with the largest deviations occurring for curved ears where the convex hull overestimates the true surface-bounded volume (\secref{sec:vol_validation}).

\subsection{Biological relevance}
Convex-hull volume and convex-hull surface area show moderate associations with manually measured ear weight ($R^{2} = 0.738$ and $0.675$, respectively; \figref{fig:trait_profiles}, panels C--D), confirming that the 3-D reconstruction captures biologically meaningful size variation. The slightly stronger association of volume relative to surface area is consistent with ear weight scaling more directly with a volumetric than a surface dimension. The residual variance reflects compounding sources: biologically, ear weight at harvest is sensitive to kernel moisture content, maturity state, and cob-to-kernel ratio differences across genotypes, none of which are recoverable from surface geometry alone; geometrically, convex-hull volume and convex-hull surface area are approximations: the convex hull overestimates the true volume for curved ears by enclosing empty space, as established in \secref{sec:vol_validation}, and the current mesh resolution does not resolve individual kernel geometry. Replacing the convex hull with a non-convex, concavity-preserving surface reconstruction would reduce the geometric component of this residual and is planned for future work. The shape descriptors (Curvature and Taper-to-Tip) are conceptually related to kernel distribution and grain-filling traits reported in the 2-D phenotyping literature \citep{Miller2017-xm,Oury2021-zt}; whether the 3-D versions of these indices provide predictive value beyond 2-D measurements is an important open question for future validation.

\subsection{Limitations}
Four constraints bound the results reported here. First, 50 of 300 ears did not reach trait extraction, and 35 of those were lost at holder segmentation rather than at reconstruction, which identifies metric scaling as the least robust stage of the pipeline. Flagged ears can be re-imaged rather than discarded, since a re-capture costs only a 20-second video and approximately one minute of operator time, so the effective loss in routine use is smaller than the 16.7\,\% first-pass figure suggests. We have not characterized the excluded ears against the retained ones, so we cannot yet rule out that attrition is correlated with ear geometry.

Second, the convex hull overestimates the volume of curved ears because it spans the concave flank, which sets a floor on the agreement achievable against water displacement. A concavity-preserving surface reconstruction would remove this bias and is the most direct route to improving volumetric accuracy.

Third, the automated quality-control gate is supplemented by visual inspection for one failure mode that the holder-eccentricity statistic does not detect, so the pipeline is automated with a single human checkpoint rather than fully unattended.

Fourth, all ears come from one diversity panel in a single season, imaged by five operators across sessions using two turntable models of the same specification. We did not isolate operator or session effects, and the Chord/Arc ratio spans only 0.965 to 1.000 in this panel, so the curvature descriptors are demonstrated over a narrow range and warrant evaluation on germplasm with more pronounced ear curvature.

\subsection{Future Directions}
Several directions are planned to address these limitations and to extend the scope of measurable traits.

\textbf{COLMAP-free pose estimation via turntable geometry.} Because the ear rotates on a motorized turntable at a known, approximately constant angular velocity, the rotation angle between successive frames provides a geometric prior on camera-pose increments. An approach under development uses per-frame rotation angles inferred from the turntable speed and frame timestamps to initialize camera poses, which are then refined jointly with the NeRF network weights through a differentiable camera-parameter optimiser in the Nerfstudio training loop. If successful, this would reduce reliance on COLMAP and potentially shorten pre-processing time, though performance relative to the current pipeline has not yet been formally evaluated. Preliminary results on a held-out subset are encouraging and will be reported in a follow-up study.

\textbf{Kernel-level trait extraction.} The trait set reported here is intentionally limited to ear-level descriptors; a dedicated pipeline for kernel-level trait extraction (including kernel row number, kernels per row, and individual kernel dimensions) builds directly on the point cloud produced by this platform in a companion study \citep{Kumar2026-tppj}, which jointly extracts eleven ear- and kernel-level traits by cylindrically unwrapping the calibrated point cloud and applying zero-shot instance segmentation, without requiring crop-specific model training. The dense color point cloud produced by the NeRF reconstruction provides a natural foundation for this extension, as full 360\textdegree{} surface color and geometry are available for localizing individual kernels without the occlusion constraints inherent to single-view approaches.
\section{Conclusions}
We have described a low-cost phenotyping pipeline that produces 3-D reconstructions of maize ears from a single 20-second video captured by a fixed consumer-grade DSLR. Combining a multi-seed COLMAP pose-estimation stage, a standard \textit{nerfacto} NeRF model, and a cylindrical-fiducial scaling procedure, the pipeline extracts skeleton length ($R^{2} = 0.964$, RMSE $= 4.68$\,mm against manual calipers), convex-hull volume ($R^{2} = 0.982$, RMSE $= 5.26$\,mL against water-displacement), and a suite of shape descriptors from the reconstructed watertight mesh. Applied to 300 ears from the SAM diversity panel, 250 (83.3\,\%) passed automated processing and quality control. Of the 50 excluded ears, 15 failed at the COLMAP registration stage and 35 were excluded at the holder-segmentation quality-control stage, so scale recovery rather than pose estimation is the dominant source of attrition. Operator involvement is approximately one minute per ear; all downstream processing runs unattended. Convex-hull volume and convex-hull surface area show moderate associations with manually measured ear weight ($R^{2} = 0.738$ and $0.675$, respectively), with residual variance attributable to both biological factors (kernel moisture content and maturity state) and the geometric limitation of the convex-hull estimator for curved ears. The trait set reported here is deliberately ear-level; a companion study \citep{Kumar2026-tppj} extends this platform to eleven ear- and kernel-level traits. COLMAP-free pose estimation via turntable-rotation priors is also under evaluation, with performance relative to the current pipeline yet to be formally established.

\section*{CRediT authorship contribution statement}
\noindent
\textbf{Therin Young:} Software, Validation, Formal analysis, Investigation, Writing -- original draft, Writing -- review \& editing, Visualization. \textbf{Elijah Rodriguez:} Methodology, Software, Validation, Visualization, Writing -- review \& editing. \textbf{Lisa Coffey:} Data curation, Writing -- review \& editing. \textbf{Talukder Zaki Jubery:} Methodology, Software, Validation, Formal analysis, Investigation, Writing -- original draft, Writing -- review \& editing, Visualization. \textbf{Adarsh Krishnamurthy:} Conceptualization, Supervision, Funding acquisition, Writing -- review \& editing. \textbf{Patrick Schnable:} Conceptualization, Resources, Supervision, Project administration, Funding acquisition, Writing -- review \& editing. \textbf{Baskar Ganapathysubramanian:} Conceptualization, Resources, Supervision, Project administration, Funding acquisition, Writing -- review \& editing.

\section*{Declaration of competing interest}
\noindent The authors declare that they have no known competing financial interests or personal relationships that could have appeared to influence the work reported in this paper.

\section*{Funding}
\noindent This work was supported by the AI Institute for Resilient Agriculture (USDA-NIFA 2021-67021-35329) and Iowa State University's Plant Science Institute. The funding sources had no involvement in study design; in the collection, analysis and interpretation of data; in the writing of the report; or in the decision to submit the article for publication.

\section*{Acknowledgments}
\noindent We thank a team of undergraduate assistants who imaged ears and generated ground truth data. Team members included: Owen Hildebrandt, Roland Stetz, Isaac Tharp, Haylee Thomas, and Ryan Wolf.

\section*{Data availability}
\noindent The source code for the reconstruction and trait-extraction pipeline is publicly available at \url{<REPOSITORY-URL>} (archived at \doi{<CODE-DOI>}). The per-ear morphological trait tables and the manual caliper, water-displacement, and ear-weight reference measurements supporting the findings of this study are deposited at \doi{<DATA DOI>}. The raw turntable videos (approximately <N>\,GB for 300 ears) are available from the corresponding authors on reasonable request owing to their size.

\section*{Declaration of generative AI and AI-assisted technologies in the manuscript preparation process}
\noindent During the preparation of this work, the authors used Claude to assist with language editing and organization. After using these tools, the authors reviewed and edited the content and take full responsibility for the final text.

\bibliographystyle{elsarticle-num-names}
\bibliography{Refs}

\newpage

\section*{Supplementary Information}

\subsubsection*{Geometry quality-control failures}

\begin{figure*}[h!]
    \centering
    \includegraphics[width=\textwidth]{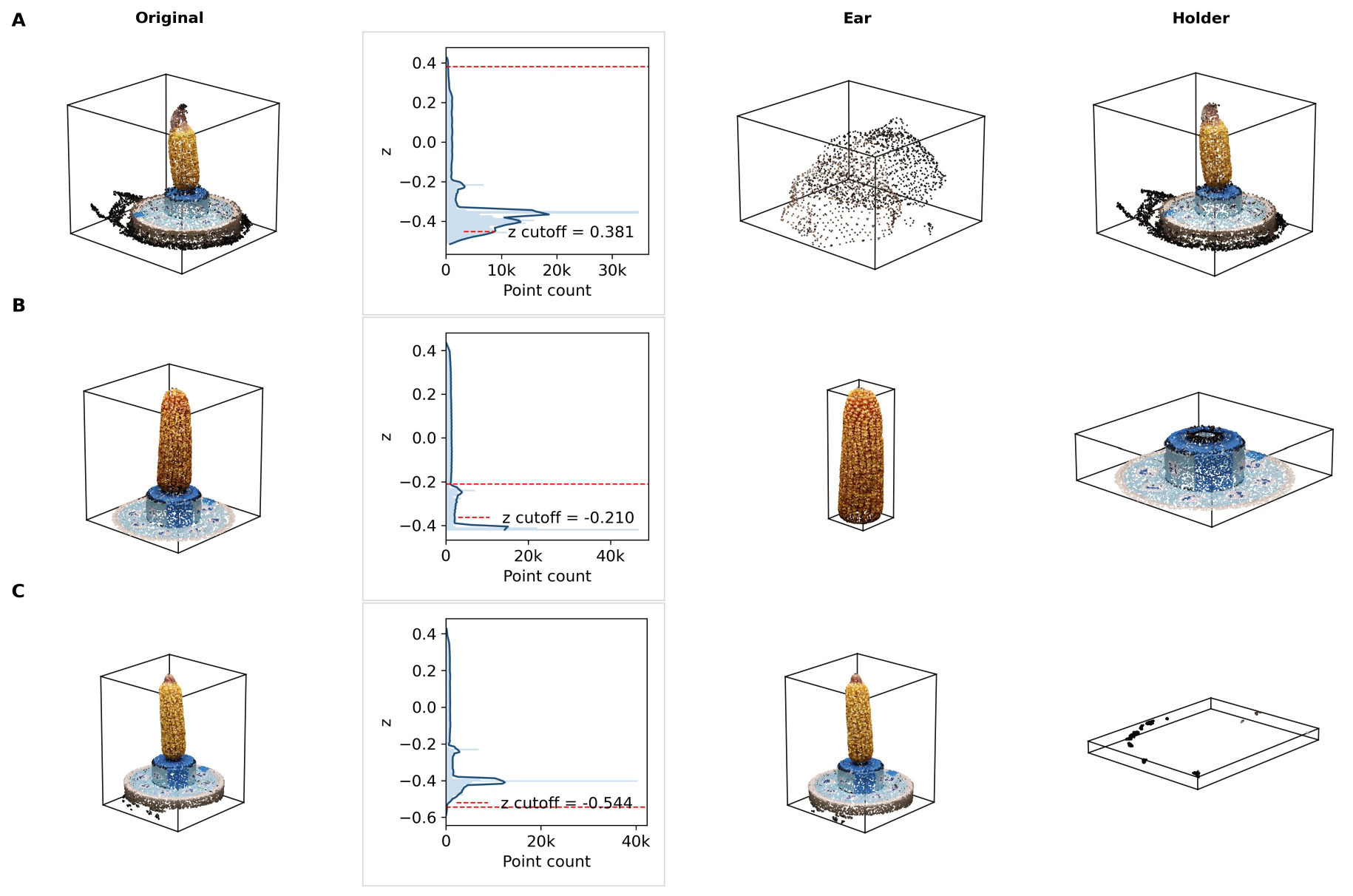}
\caption{Each column illustrates a distinct failure mode.
\textbf{(A)}~Partial ear detection: the cutoff boundary underestimates the holder--ear junction, causing the upper portion of the ear to be excluded from the segmented region. The holder cross-section is approximately circular but the ear mesh passed to trait extraction is incomplete. This case was identified during manual inspection, as $e_{\text{holder}}$ alone does not flag it.
\textbf{(B)}~Contaminated segmentation: the cutoff boundary falls within the ear body rather than at the holder rim, so structural parts of the ear are included in the region fitted as the holder. The resulting cross-section is non-circular ($e_{\text{holder}}$ elevated above the IQR fence) and the derived scale factor is unreliable.
\textbf{(C)}~Holder mis-identification: the isolated region does not correspond to the cylindrical holder, yielding a fitted radius $\hat{r}$ that falls well outside the expected session range. The circle fit produces an erroneous scale factor that would propagate into all physical-unit traits if not caught.
Failure modes (B) and (C) are detected automatically by the eccentricity and radius checks respectively (\crefs{Eqs.}{eq:eholder}{eq:iqr}). Failure mode (A) requires manual review and motivates visual inspection of the segmented point cloud as a routine quality-control step. The underlying cause in each case (a flat $z$-density profile, partial holder occlusion, or an ear positioned too close to the holder rim) is noted below the respective panel. }
    \label{fig:qc_failures}
\end{figure*}

\figref{fig:qc_failures} shows the reconstructed holder cross-sections for a representative sample of ears excluded by the geometry quality-control filter described in \secref{sec:qc}.

\end{document}